\documentclass{article}
\PassOptionsToPackage{numbers, compress}{natbib}

\usepackage[preprint]{neurips_2026}

\usepackage[utf8]{inputenc} 
\usepackage[T1]{fontenc}    
\usepackage{hyperref}       
\usepackage{url}            
\usepackage{booktabs}       
\usepackage{amsfonts}       
\usepackage{nicefrac}       
\usepackage{microtype}      
\usepackage{xcolor}         

\usepackage{microtype}
\usepackage{graphicx}
\usepackage{subcaption}
\usepackage{booktabs} 
\usepackage{pifont} 

\usepackage{amssymb}
\usepackage{amsfonts}       
\usepackage{nicefrac}       
\usepackage{microtype}      
\usepackage{xcolor}         
\usepackage{amsmath}        
\usepackage{amsthm}         
\usepackage{resizegather}   
\usepackage{graphicx}       
\usepackage{pifont}         
\usepackage{float}

\usepackage{algorithm}
\usepackage{algorithmic}
\usepackage{adjustbox}
\usepackage{graphicx}
\usepackage{multirow}
\usepackage{adjustbox}
\usepackage{tabularx}
\usepackage{booktabs}
\usepackage{caption}
\usepackage{hyperref}

\AtBeginDocument{
\setlength{\abovedisplayskip}{5pt}
\setlength{\belowdisplayskip}{5pt}
}
\usepackage[font=small,labelfont=bf]{caption}

\usepackage{amsmath}
\usepackage{amssymb}
\usepackage{mathtools}
\usepackage{amsthm}

\title{DeepLévy: Learning Heavy-Tailed Uncertainty in  \\ Highly Volatile Time Series}

\author{%
  Yang Yang\\
  University of New South Wales\\
  Australia NSW Sydney\\
  \texttt{yang.yang26@unsw.edu.a} \\
    \And
    Du Yin \\
  University of New South Wales \\
  Sydney NSW 2052 Australia \\
   \texttt{du.yin@unsw.edu.au} \\
  \AND
  Hao Xue \\
  The Hong Kong University of \\ Science and Technology (Guangzhou) \\
  \texttt{haoxue@hkust-gz.edu.cn} \\
  \And
  Flora Salim \\
  University of New South Wales \\
  Sydney NSW 2052 Australia \\
  \texttt{flora.salim@unsw.edu.au} \\
}

\begin{document}

\maketitle

\begin{abstract}
Modeling uncertainty in heavy-tailed time series remains a critical challenge for deep probabilistic forecasting models, which often struggle to capture abrupt, extreme events. While Lévy stable distributions offer a natural framework for modeling such non-Gaussian behaviors, the intractability of their probability density functions severely limits conventional likelihood-based inference. To address this, we introduce DeepLévy, a neural framework that learns mixtures of Lévy stable distributions by minimizing the discrepancy between empirical and parametric characteristic functions. DeepLévy incorporates a mixture mechanism that adaptively learns context-dependent weights and parameters over multiple Lévy components, enabling flexible multi-horizon uncertainty modeling. Evaluations on both real and synthetic datasets demonstrate that DeepLévy outperforms state-of-the-art deep probabilistic forecasting approaches in tail risk metrics, especially under extreme volatility.
\end{abstract}

\section{Introduction}

\begin{figure}[!th]
\centering
\centerline{\includegraphics[width=1\columnwidth]{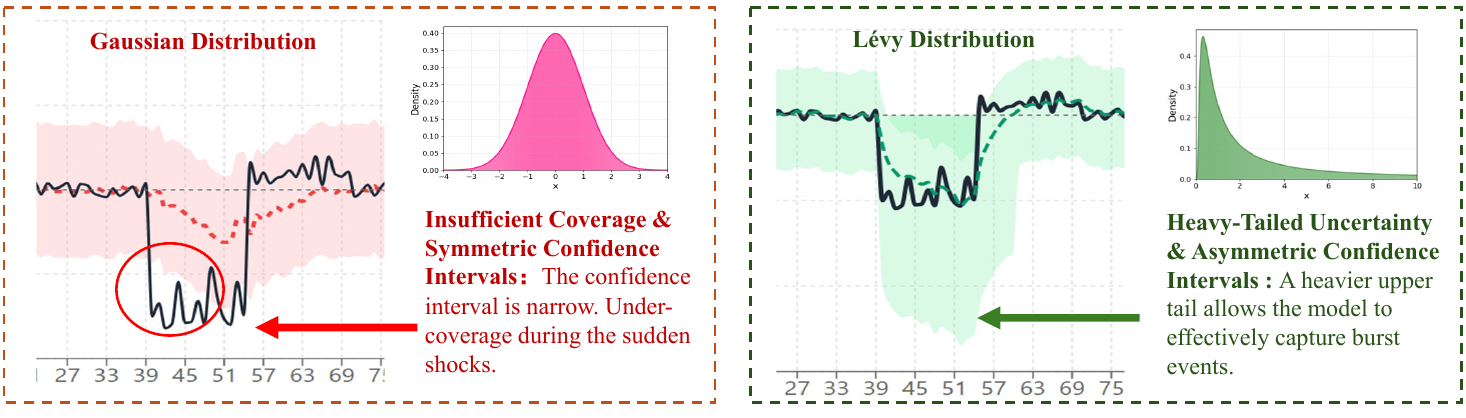}} 
\caption{Gaussian vs. Lévy distributions in extreme events. The Gaussian assumption (Top) underestimates tail risk due to symmetry, whereas the Lévy distribution (Bottom) captures the skewness and heavy-tailed nature of sudden shocks.}
\vspace{-13pt}
\label{fig:contrast}
\end{figure}

Forecasting extreme events in time series plays a pivotal role in high-stakes domains such as epidemiological crisis management~\citep{cao2024covid,salim2021learning} and financial risk assessment~\citep{thomas2000survey}. Given a historical sequence $\mathbf{x}_{1:T}$, conventional deep forecasting methods typically train a model $f(\mathbf{x}_{1:T})$ to approximate the conditional distribution $P(\mathbf{y}_{T+1:T+H}|\mathbf{x}_{1:T})$ over a prediction horizon $H$. Although recent advancements in deep learning have demonstrated remarkable capabilities in modeling temporal dependencies~\citep{girish2018crop,miller2024survey}, effective decision-making, particularly during "black swan" events like market crashes or pandemic surges, requires accurately estimating the tail risks underlying the data~\citep{rehman2024black}. To address this, many recent studies have pivoted towards probabilistic forecasting~\citep{gneiting2014probabilistic,di2023explainable}, aiming to quantify the full range of possible future outcomes rather than merely predicting a deterministic trajectory. Deep neural-based probabilistic forecasting models, such as DeepAR~\citep{salinas2020deepar}, DeepVAR~\citep{salinas2019high}, DSSM~\citep{rangapuram2018deep}, DeepFactor~\citep{wang2019deep}, ProbTransformer~\citep{tang2021probabilistic}, etc., have gained significant traction due to their ability to output valid probability density functions. However, existing methods predominantly rely on the assumption that the noise $\boldsymbol{\epsilon}$ follows a Gaussian or light-tailed distribution with finite variance~\cite{jorgensen2012statistical}. While effective for routine data, this assumption fundamentally fails in heavy-tailed regimes where the variance may be infinite and the probability of extreme deviations decays polynomially rather than exponentially~\citep{resnick2007heavy,rocco2014extreme}. The primary objective of these models is restricted by the expressiveness of standard parametric families, which cannot capture the infinite variance and extreme asymmetry inherent in complex real-world systems. As illustrated in Figure~\ref{fig:contrast}, when applied to data with power-law tails, Gaussian-based models systematically underestimate the probability of catastrophic events, leading to critical failures in risk assessment.

To better address the challenges of heavy-tailed and multimodal dynamics, we introduce the \textbf{DeepLévy}, a novel framework that fundamentally relaxes the finite-variance assumption by modeling the conditional distribution as a mixture of Lévy $\alpha$-stable distributions~\citep{barthelemy2008levy,tsallis1997levy}. Unlike Gaussian distributions, Lévy stable distributions are characterized by a stability index $\alpha \in (0, 2]$, a skewness parameter $\beta \in [-1, 1]$, a scale parameter $\gamma > 0$, and a location parameter $\delta \in \mathbb{R}$, allowing them to model varying degrees of tail heaviness and asymmetry. However, the direct application of stable distributions in deep learning is hindered by the lack of closed-form probability density functions \citep{nolan2020univariate, nolan2022computational}. We overcome this obstacle by operating in the spectral domain: since stable distributions admit known characteristic functions (CFs), we train DeepLévy by matching the predicted CF to the empirical CF of the targets~\citep{koutrouvelis1980regression,kogon1998characteristic}. To address the numerical challenges of CF-based learning, including the rapid decay of the signal-to-noise ratio at high frequencies, we introduce a scale-adaptive frequency weighting scheme and use the continuous $S_0$-parameterization for gradient stability. In summary, our contributions are as follows:

\begin{itemize}
    \item We identify that standard finite-variance assumptions are inadequate for modeling extreme events and propose a novel neural architecture that integrates a mixture of Lévy stable distributions for multi-horizon forecasting. This is among the first works to effectively combine deep learning with the full parameter space of stable distributions ($\alpha$, $\beta$, $\gamma$, $\delta$) for time series forecasting.
    
    \item To resolve the numerical instability and gradient explosion issues inherent in stable distribution parameterization, we propose a robust optimization strategy involving: (i) a scale-adaptive frequency weighting scheme that dynamically adapts to the predicted distribution's decay rate, (ii) the continuous $S_0$-parameterization that eliminates discontinuities at $\alpha = 1$, and (iii) constrained parameter projections that maintain mathematical validity throughout training.
    
    \item Experimental results demonstrate that DeepLévy achieves superior performance in capturing tail risks. Specifically, compared to state-of-the-art probabilistic baselines, our model significantly improves the accuracy of tail estimation on both synthetic heavy-tailed datasets and real-world financial and epidemiological benchmarks.
\end{itemize}

\section{Related Works}

\textbf{Deep Probabilistic Forecasting:} Deep probabilistic forecasting estimates the full predictive distribution rather than point predictions alone~\citep{gneiting2014probabilistic}. Standard parametric approaches, including RNN-based models like DeepAR~\citep{salinas2020deepar}, DeepVAR~\citep{salinas2019high}, and DSSM~\citep{rangapuram2018deep}, alongside recent Transformer-based or causal architectures such as ProbTransformer~\citep{tang2021probabilistic}, MTSNet~\citep{yang2023mtsnet}, and STOAT~\citep{yang2025stoat}, typically combine neural sequence encoders with parametric likelihood heads. While these methods have explored a spectrum of distributions ranging from Gaussian and Laplace to Student's $t$, they fundamentally rely on the existence of a closed-form probability density function (PDF) to perform exact maximum likelihood estimation (MLE). This dependence constitutes a prohibitive barrier for modeling Lévy stable distributions, which generally lack analytical PDFs. Although recent studies have proposed approximating the cumulative distribution function (CDF) of Lévy processes using mixtures of logistic distributions for option pricing simulations~\citep{kudryavtsev2023applications}, such methods are tailored for Monte Carlo generation rather than direct multi-horizon forecasting. Consequently, effective modeling of heavy-tailed dynamics in sequence forecasting necessitates alternative estimation strategies beyond classical MLE and CDF approximations. 

\textbf{Heavy-Tailed Modeling and Lévy Assumption:} To address the limitations of Gaussian assumptions in modeling extreme events, parametric approaches often employ the Student's $t$ distribution due to its tractable density~\citep{lin2006can}. However, its reliance on symmetric tails and a single degrees-of-freedom parameter restricts its ability to capture the directional asymmetry and time-varying tail dynamics characteristic of financial crashes or epidemic outbreaks. Lévy distributions~\cite{tsallis1997levy} offer a rigorous generalization, employing separate parameters for tail heaviness and skewness to naturally model infinite variance and asymmetric extremes. Despite their theoretical advantages, estimating stable parameters is notoriously difficult due to the lack of closed-form PDFs. Classical characteristic function estimators~\citep{koutrouvelis1980regression} and Bayesian MCMC approaches~\citep{cauchemez2004bayesian} are often computationally prohibitive for high-dimensional gradient-based learning. While recent neural approaches like \citet{xu2020calibrating} also exploit characteristic functions to sidestep intractable multivariate Lévy densities, their objective is inverse modeling: calibrating a Lévy process to match observed increments under a stationary/incremental likelihood surrogate. DeepL\'{e}vy instead utilizes CF matching as a forward training signal for a sequence-to-sequence forecaster. In this framework, an encoder--decoder predicts horizon- and context-dependent mixture stable parameters, while the loss function compares predicted mixture CFs to single-horizon empirical CF exponentials using a scale-adaptive frequency weighting scheme. Consequently, the spectral-domain intuition is related, yet DeepL\'{e}vy differs substantially in its focus on conditional multi-horizon forecasting rather than stationary process calibration, its use of context-varying mixtures, and its unique optimization template.

\vspace{-0.2cm}
\section{Preliminary}

\textbf{Lévy $\alpha$-Stable Distributions.} 
The Lévy $\alpha$-stable distribution, denoted as $\mathcal{S}(\alpha, \beta, \gamma, \delta)$, is a class of heavy-tailed probability distributions that generalizes the Gaussian family. Unlike Gaussian distributions, general stable distributions do not possess closed-form probability density functions (PDFs), except for specific cases. Consequently, they are canonically defined by their characteristic function. To ensure numerical stability and continuity of the parameters, particularly near $\alpha=1$, we adopt Nolan's $S_0$-parameterization (the ``$0$'' form in~\citet{nolan2020univariate}). The characteristic function is
\begin{subequations}
\label{eq:cf_s0_all}
\begin{equation}
\label{eq:cf_s0}
    \varphi(\tau; \alpha, \beta, \gamma, \delta) = 
        \exp\Biggl(-|\gamma\tau|^\alpha \Biggl[1 + i\beta\,\text{sgn}(\tau)\tan\frac{\pi\alpha}{2}\left(|\gamma\tau|^{1-\alpha} - 1\right)\Biggr] + i\delta\tau\Biggr), \quad \alpha \neq 1,
\end{equation}
\begin{equation}
\label{eq:cf_s0_alpha1}
    \varphi(\tau; 1, \beta, \gamma, \delta) = 
        \exp\left(-|\gamma\tau| \left[1 + i\beta\,\text{sgn}(\tau)\frac{2}{\pi}\ln|\gamma\tau|\right] + i\delta\tau\right).
\end{equation}
\end{subequations}
Here, $\alpha \in (0, 2]$ is the stability index governing tail thickness (smaller $\alpha$ indicates heavier tails), $\beta \in [-1, 1]$ controls skewness, $\gamma > 0$ is the scale parameter, and $\delta \in \mathbb{R}$ is the location parameter. The factor $\left(|\gamma\tau|^{1-\alpha} - 1\right)$ in~\eqref{eq:cf_s0} cancels the singular growth of $\tan(\pi\alpha/2)$ as $\alpha \to 1$, so~\eqref{eq:cf_s0} varies continuously in $\alpha$ and joins~\eqref{eq:cf_s0_alpha1} at $\alpha=1$. This should not be confused with the classical $S_1$ form $\exp\bigl(-|\gamma\tau|^\alpha[1 - i\beta\,\text{sgn}(\tau)\tan(\pi\alpha/2)] + i\delta\tau\bigr)$, which is discontinuous in $\alpha$ at $\alpha=1$ when $\beta \neq 0$.

\subsection{Problem Formulation}

\textbf{General Probabilistic Forecasting.}
Let $\mathbf{x}_{1:T} = (x_1, x_2, \ldots, x_T) \in \mathbb{R}^{T}$ denote the historical observations over a context window of length $T$. The goal of probabilistic forecasting is to estimate the joint conditional probability distribution $\mathbb{P}(\mathbf{y}_{T+1:T+H} | \mathbf{x}_{1:T})$ for the future trajectory $\mathbf{y}_{T+1:T+H} = (y_{T+1}, \ldots, y_{T+H}) \in \mathbb{R}^{H}$ over a prediction horizon $H \geq 1$.

\textbf{Limitations of Gaussian and Light-Tailed Assumptions.}
State-of-the-art deep probabilistic forecasting methods, such as DeepAR~\citep{salinas2020deepar}, typically assume the conditional distribution belongs to a parametric family with exponentially decaying tails. These methods learn a mapping $f_\theta: \mathbb{R}^{T} \to \mathcal{P}^H$ from historical observations to distribution parameters for each horizon, where $\mathcal{P}$ denotes the parameter space of the assumed distribution family. The model is optimized via maximum likelihood:
\begin{equation}
    \max_\theta \mathbb{E} \left[ \sum_{h=1}^{H} \log p(y_{T+h}; f_\theta^{(h)}(\mathbf{x}_{1:T}, \mathbf{y}_{T+1:T+h-1})) \right],
\end{equation}
where $p(y; \cdot)$ denotes the probability density function parameterized by the network output, and $f_\theta^{(h)}$ denotes the parameters for horizon $h$. This formulation is fundamentally limited when the underlying process exhibits (1) \textbf{power-law tails} with potentially infinite variance, and (2) \textbf{extreme asymmetry} beyond what light-tailed distributions can capture.

\textbf{Proposed Formulation: Heavy-Tailed Mixture via Spectral Matching.}
We hypothesize that the conditional distribution at each horizon $h$ follows a Mixture of Lévy Stable Distributions with $K$ components:
\begin{equation}
\label{eq:mixture}
    p(y_{T+h} | \mathbf{x}_{1:T}, \mathbf{y}_{T+1:T+h-1}) = \sum_{k=1}^K \pi_k^{(h)} \cdot p_{\mathcal{S}}(y_{T+h}; \alpha_k^{(h)}, \beta_k^{(h)}, \gamma_k^{(h)}, \delta_k^{(h)}),
\end{equation}
where $p_{\mathcal{S}}(y; \alpha, \beta, \gamma, \delta)$ denotes the probability density of a Lévy stable distribution $\mathcal{S}(\alpha, \beta, \gamma, \delta)$. For each component $k \in \{1, \ldots, K\}$ at horizon $h$: $\pi_k^{(h)} \in [0,1]$ is the mixing weight satisfying $\sum_{k=1}^K \pi_k^{(h)} = 1$; $\alpha_k^{(h)} \in (0, 2]$ is the stability index controlling tail heaviness (smaller $\alpha$ implies heavier tails); $\beta_k^{(h)} \in [-1, 1]$ is the skewness parameter; $\gamma_k^{(h)} > 0$ is the scale parameter; and $\delta_k^{(h)} \in \mathbb{R}$ is the location parameter. All mixture parameters are context-dependent and predicted by a neural network $\mathcal{G}_\Theta$ with parameters $\Theta$:
\begin{equation}
\label{eq:network}
    \{\pi_k^{(h)}, \alpha_k^{(h)}, \beta_k^{(h)}, \gamma_k^{(h)}, \delta_k^{(h)}\}_{k=1}^K = \mathcal{G}_\Theta^{(h)}(\mathbf{x}_{1:T}, \mathbf{y}_{T+1:T+h-1}).
\end{equation}
Since stable distributions lack closed-form densities (except for special cases), we operate in the spectral domain via characteristic functions. Let $i = \sqrt{-1}$ denote the imaginary unit. The learning objective minimizes the weighted squared distance between the predicted mixture characteristic function (CF) and the empirical CF over a discrete frequency grid $\mathcal{T} = \{\tau_1, \ldots, \tau_M\}$ with $M$ evaluation points:
\begin{equation}
\label{eq:loss_formulation}
    \mathcal{L}(\Theta) = \mathbb{E} \left[ \sum_{h=1}^{H} \frac{1}{W^{(h)}} \sum_{m=1}^M w^{(h)}(\tau_m) \left| \Phi_{\text{mix}}^{(h)}(\tau_m) - e^{i \tau_m y_{T+h}} \right|^2 \right],
\end{equation}
where $\Phi_{\text{mix}}^{(h)}(\tau) = \sum_{k=1}^K \pi_k^{(h)} \varphi_k^{(h)}(\tau)$ is the mixture CF at horizon $h$ with $\varphi_k^{(h)}(\tau)$ being the CF of the $k$-th stable component, $W^{(h)} = \sum_{m=1}^M w^{(h)}(\tau_m)$ is the normalization constant, and $w^{(h)}(\tau) > 0$ is a sample-dependent weighting function designed to match the CF decay rate (detailed in Section~\ref{sec:methodology}).
\vspace{-0.2cm}
\section{Methodology}
\label{sec:methodology}
\vspace{-0.2cm}

\begin{figure}[!th]
\centering
\includegraphics[width=1\linewidth]{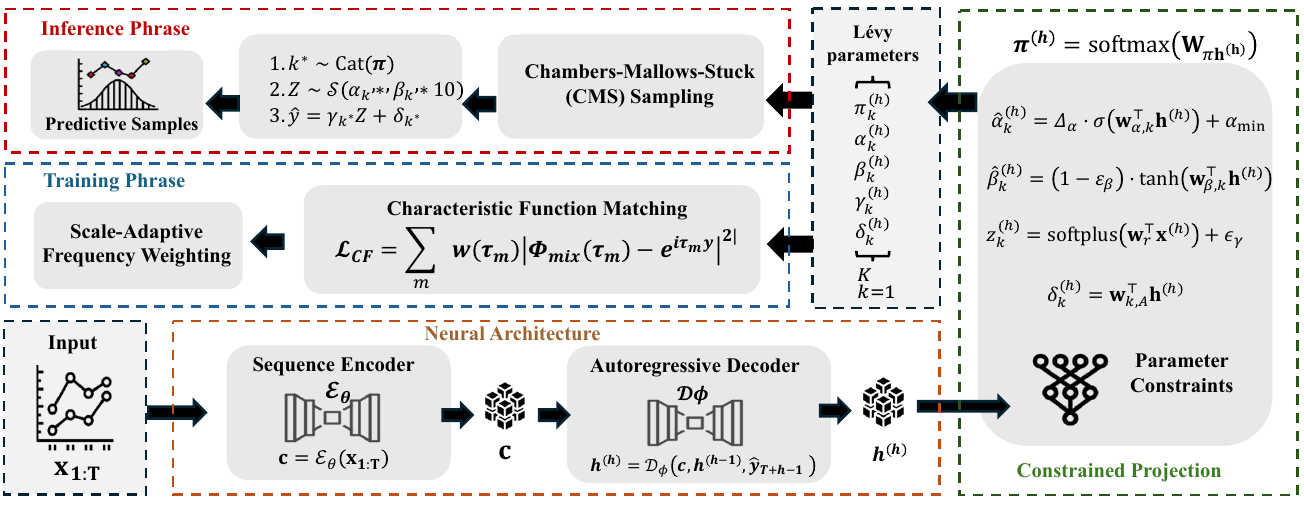}
\caption{DeepLévy integrates a neural sequence encoder with an autoregressive decoder and constrained mixture projection layer for multi-horizon forecasting. During the training phase, a fixed frequency grid with scale-adaptive weighting and a variance-reduced characteristic function loss are utilized to optimize the distribution parameters without explicit density estimation.}
\label{fig:overview}
\vspace{-0.3cm}
\end{figure}

In this section, we present DeepLévy, our framework for heavy-tailed probabilistic forecasting. The architecture follows a modular design: (1) a sequence encoder maps historical observations to a latent representation, (2) an autoregressive decoder generates horizon-specific representations, (3) a constrained projection layer transforms these representations into valid mixture parameters, (4) characteristic function matching is performed with scale-adaptive weighting for training, and (5) the Chambers-Mallows-Stuck algorithm generates samples during inference. The outline of DeepLévy is given in Figure~\ref{fig:overview}. Given a historical sequence $\mathbf{x}_{1:T} \in \mathbb{R}^{T}$, our goal is to predict the parameters of a $K$-component Lévy stable mixture distribution for each future horizon $h \in \{1, \ldots, H\}$. The architecture consists of an encoder-decoder structure that enables multi-horizon forecasting.

\textbf{Encoder.} The encoder maps the raw historical observations to a context-aware latent representation $\mathbf{c} \in \mathbb{R}^d$ via a sequence encoder $\mathcal{E}_\theta$:
\begin{equation}
\label{eq:encoder}
    \mathbf{c} = \mathcal{E}_\theta(\mathbf{x}_{1:T}),
\end{equation}
where $d$ is the hidden dimension, and $\theta \subset \Theta$ denotes the encoder parameters. The encoder $\mathcal{E}_\theta$ can be parameterized by architectures such as Transformers~\citep{vaswani2017attention}, or Long Short-Term Memory (LSTM) networks~\citep{hochreiter1997long}.

\textbf{Autoregressive Decoder.} To capture dependencies across prediction horizons, we employ an autoregressive decoder that generates horizon-specific hidden states:
\begin{equation}
\label{eq:decoder}
    \mathbf{h}^{(h)} = \mathcal{D}_\phi(\mathbf{c}, \mathbf{h}^{(h-1)}, \hat{y}_{T+h-1}), \quad h = 1, \ldots, H,
\end{equation}
where $\mathbf{h}^{(0)} = \mathbf{c}$, $\hat{y}_{T}$ is initialized as the last observation $x_T$, and $\hat{y}_{T+h-1}$ denotes the autoregressive feedback from horizon $(h-1)$. During training we use teacher forcing with the realized targets $y_{T+h-1}$ whenever available; when simulating rollouts from the model, $\hat{y}_{T+h-1}$ is a sample from the predicted mixture at $(h-1)$. We do not use the conditional mean as feedback when $\alpha \le 1$, because it need not exist for stable laws in that regime; any optional point summary for analysis uses the location parameter $\delta$ (which remains well defined under $S_0$), not the mean. The decoder $\mathcal{D}_\phi$ can be implemented as an LSTM cell or Transformer decoder layer. The latent vector $\mathbf{h}^{(h)}$ serves as a sufficient statistic summarizing both historical context and autoregressive information for horizon $h$. The next step is to decode this representation into valid distributional parameters.

\vspace{-0.1cm}

\subsection{Constrained Parameter Projection}
\label{sec:projection}

To transform the latent representation $\mathbf{h}^{(h)}$ into valid mixture parameters $\{\pi_k^{(h)}, \alpha_k^{(h)}, \beta_k^{(h)}, \gamma_k^{(h)}, \delta_k^{(h)}\}_{k=1}^K$, we employ a constrained projection mechanism. Unlike Gaussian mixtures, Lévy stable distributions require strict constraints to ensure mathematical validity and numerical stability during the computation of the characteristic function. We use the hat notation (e.g., $\hat{\alpha}_k^{(h)}$) to denote the network-predicted parameters, distinguishing them from the generic mathematical symbols used in distribution definitions.

\textbf{Mixing Weights.} The mixing coefficients $\boldsymbol{\pi}^{(h)} = (\pi_1^{(h)}, \ldots, \pi_K^{(h)})^\top \in \mathbb{R}^K$ must satisfy the simplex constraint $\sum_{k=1}^K \pi_k^{(h)} = 1$ with $\pi_k^{(h)} \geq 0$. We enforce this via a softmax transformation:
\begin{equation}
\label{eq:mixing_weights}
    \boldsymbol{\pi}^{(h)} = \text{softmax}(\mathbf{W}_\pi \mathbf{h}^{(h)}),
\end{equation}
where $\mathbf{W}_\pi \in \mathbb{R}^{K \times d}$ is a learnable weight matrix.

\textbf{Component Parameters.} For each mixture component $k \in \{1, \ldots, K\}$, the four L\'{e}vy stable parameters are obtained through constrained projections from the latent representation $\mathbf{h}^{(h)}$:
\begin{equation}
\small
\label{eq:param_constraints}
\begin{aligned}
    \hat{\alpha}_k^{(h)} &= \Delta_\alpha \cdot \sigma(\mathbf{w}_{\alpha,k}^\top \mathbf{h}^{(h)}) + \alpha_{\min}, \\
    \hat{\beta}_k^{(h)} &= (1-\epsilon_{\beta}^{\max}) \cdot \tanh(\mathbf{w}_{\beta,k}^\top \mathbf{h}^{(h)}), \\
    \hat{\gamma}_k^{(h)} &= \mathrm{softplus}(\mathbf{w}_{\gamma,k}^\top \mathbf{h}^{(h)}) + \epsilon_\gamma, \\
    \hat{\delta}_k^{(h)} &= \mathbf{w}_{\delta,k}^\top \mathbf{h}^{(h)},
\end{aligned}
\end{equation}
where $\sigma(\cdot)$ denotes the sigmoid function, $\alpha_{\min}=0.1$, $\alpha_{\max}=1.95$, and $\Delta_\alpha=\alpha_{\max}-\alpha_{\min}=1.85$. We set $\epsilon_{\beta}^{\max}=0.02$, yielding $\hat{\beta}_k^{(h)} \in (-0.98,0.98)$, and $\epsilon_\gamma = 10^{-4}$ to prevent numerical underflow. The vectors $\{\mathbf{w}_{\alpha,k}, \mathbf{w}_{\beta,k}, \mathbf{w}_{\gamma,k}, \mathbf{w}_{\delta,k}\} \in \mathbb{R}^d$ are learnable projection vectors. The stability index $\hat{\alpha}_k^{(h)} \in [0.1, 1.95]$ controls tail heaviness, the skewness parameter governs asymmetry, the scale parameter $\hat{\gamma}_k^{(h)} \in (\epsilon_\gamma, +\infty)$ determines dispersion, and the location parameter $\hat{\delta}_k^{(h)} \in \mathbb{R}$ specifies the center. These constrained ranges are chosen to balance expressiveness with numerical stability during characteristic function computation. More detailed discussions can be found in Appendix~\ref{appendix:numerical_stability}.

\subsection{Characteristic Function Matching}
\label{sec:cf_matching}
Since the probability density function of a general Lévy stable distribution lacks a closed form~\citep{nolan2020univariate,samorodnitsky1994stable}, we train the model by matching the predicted characteristic function to the empirical characteristic function of the targets. This approach, known as the empirical characteristic function (ECF) method, has been widely used for stable distribution estimation~\citep{koutrouvelis1980regression,kogon1998characteristic,bruck2025generative,yu2024cf}.

\textbf{Frequency Grid and Characteristic Function Computation.}The characteristic function is theoretically defined over all frequencies $\tau \in \mathbb{R}$. For practical computation, we construct a fixed symmetric frequency grid $\mathcal{T}$ as a uniform linspace over $[-\tau_{\max}, \tau_{\max}]$ with $M$ points. For standardized input data, we set $\tau_{\max} = 15.0$ to provide sufficient coverage of the informative frequency range. We compute the characteristic function using the $S_0$ parameterization in~\eqref{eq:cf_s0}--\eqref{eq:cf_s0_alpha1}. To ensure numerical stability, we work with the log-characteristic function $\psi(\tau) = \ln\varphi(\tau)$, decomposed into real and imaginary parts; although $S_0$ is analytically continuous in $\alpha$ at $\alpha=1$, we evaluate the skew term $\tan(\pi\alpha/2)\bigl(|\gamma\tau|^{1-\alpha}-1\bigr)$ in a numerically stable way and optionally switch to the closed $\alpha=1$ expression in a narrow band for speed. The component CF is computed using Euler's formula, and the mixture CF is obtained by weighted summation:
\begin{equation}
\label{eq:cf_mixture}
    \Phi_{\text{mix}}^{(h)}(\tau) = \sum_{k=1}^K \pi_k^{(h)} \varphi_k^{(h)}(\tau).
\end{equation}
The detailed procedure for numerically stable CF computation under the $S_0$ parameterization is provided in Algorithm~\ref{alg:cf_computation} of Appendix~\ref{appendix:cf_computation}.

\textbf{Scale-Adaptive Frequency Weighting.} A key challenge in CF matching is that the signal-to-noise ratio (SNR) of the empirical CF decays rapidly with frequency \citep{nolan2020univariate}. Uniformly weighting all frequencies leads to overfitting on high-frequency noise. We address this by applying a scale-adaptive weighting function that matches the expected CF decay rate.

Given the predicted parameters, we first compute the effective scale and stability parameters as mixture-weighted averages:
\begin{equation}
\label{eq:effective_params}
    \gamma_{\text{eff}}^{(h)} = \sum_{k=1}^K \pi_k^{(h)} \hat{\gamma}_k^{(h)}, \quad
    \alpha_{\text{eff}}^{(h)} = \sum_{k=1}^K \pi_k^{(h)} \hat{\alpha}_k^{(h)}.
\end{equation}
The weighting function is then defined as:
\begin{equation}
\label{eq:weighting_function}
    w^{(h)}(\tau) = \exp\left(-\left|\text{sg}(\gamma_{\text{eff}}^{(h)}) \tau\right|^{\text{sg}(\alpha_{\text{eff}}^{(h)})}\right),
\end{equation}
where $\text{sg}(\cdot)$ denotes the stop-gradient operator, which prevents the moving target problem during optimization.

\textbf{Loss Formulation.} The empirical characteristic function for a target observation $y_{T+h}$ is simply the complex exponential $\phi_{\text{emp}}^{(h)}(\tau) = e^{i\tau y_{T+h}}$. The CF loss is formulated as the weighted mean squared error between the predicted mixture CF and the empirical CF, aggregated over all horizons. For a minibatch of $B$ samples:
\begin{equation}
\label{eq:cf_loss}
    \mathcal{L}_{\text{CF}} = \frac{1}{B} \sum_{b=1}^B \sum_{h=1}^{H} \frac{\sum_{m: |\tau_m| > \epsilon_\tau} w^{(h,b)}(\tau_m) \left|\Phi_{\text{mix}}^{(h,b)}(\tau_m) - \phi_{\text{emp}}^{(h,b)}(\tau_m)\right|^2}{\left(\sum_{m: |\tau_m| > \epsilon_\tau} w^{(h,b)}(\tau_m)\right) + \epsilon_W},
\end{equation}
where $\epsilon_\tau = 10^{-12}$ excludes near-zero frequencies, and $\epsilon_W = 10^{-8}$ prevents division by zero. Since the empirical CF based on a single observation has high variance, we leverage batch aggregation to reduce gradient variance. For each frequency $\tau_m$, the batch provides $B$ independent estimates of the CF discrepancy, and averaging over the batch reduces the variance by a factor of $B$. This is why we require sufficiently large batch sizes ($B \geq 128$) for stable training. Our CF objective admits a standard identification interpretation from the classical uniqueness of characteristic functions: under population-level expectations and sufficient frequency support, the weighted CF discrepancy is minimized if and only if the model and target conditional distributions coincide. This is used as a soundness argument rather than a new standalone theory contribution. We provide a formal proposition and a finite-batch moment lemma for empirical CF estimation in Appendix~\ref{appendix:cf_theory}; the lemma also explains why larger mini-batches reduce optimization noise.

\textbf{Entropy Regularization.} To encourage the model to utilize multiple mixture components and prevent mode collapse~\citep{pereyra2017regularizing}, we add an entropy regularization term on the mixing weights and the $\lambda_{\text{ent}} \geq 0$ controls the regularization strength:
{\small
\begin{equation}
\label{eq:combined_loss}
\begin{aligned}
\mathcal{L}_{\text{total}} &= \mathcal{L}_{\text{CF}} - \lambda_{\text{ent}} H\\
H &= \frac{1}{BH} \sum_{b=1}^B \sum_{h=1}^{H} \left(-\sum_{k=1}^K \pi_k^{(h,b)} \ln(\pi_k^{(h,b)}+\epsilon_H)\right)
\end{aligned}
\end{equation}
}

\textbf{Gradient Clipping.} Heavy-tailed distributions can produce large gradient magnitudes. To ensure stable training, we apply Gradient Clipping~\citep{pascanu2013difficulty} with the parameter update rule:
\begin{equation}
\label{eq:gradient_update}
    \Theta \leftarrow \Theta - \eta \cdot \frac{\mathbf{g}}{\max(1, \|\mathbf{g}\|_2)},
\end{equation}
where $\mathbf{g} = \nabla_\Theta \mathcal{L}_{\text{total}}$ and $\eta > 0$ is the learning rate. A complete summary of all numerical constants used in DeepLévy is provided in Appendix~\ref{appendix:numerical_constants}.

\subsection{Inference and Sampling}
\label{sec:inference}

During inference, the goal is to generate predictive samples from the learned mixture distribution over the entire prediction horizon. Given a new historical sequence $\mathbf{x}_{1:T}$, we first compute the context representation $\mathbf{c} = \mathcal{E}_\theta(\mathbf{x}_{1:T})$ and then autoregressively generate samples for each horizon. For each horizon $h = 1, \ldots, H$: (1) compute the horizon-specific hidden state $\mathbf{h}^{(h)}$ using Eq.~\eqref{eq:decoder}; (2) obtain the predicted parameters via the constrained projection; (3) sample a component index $k^* \sim \text{Categorical}(\boldsymbol{\pi}^{(h)})$ and generate a sample from that component using the Chambers-Mallows-Stuck (CMS) algorithm~\citep{chambers1976method} (see Algorithm~\ref{alg:cms_sampling} for the complete procedure under $S_0$ parameterization); (4) feed the sampled value back to the decoder for the next horizon. The CMS algorithm enables efficient and mathematically rigorous sampling from the heavy-tailed predictive distribution under the $S_0$ parameterization. For probabilistic forecasting, we generate $N_{\text{samples}}$ independent trajectories by repeating the entire autoregressive sampling process, which provides Monte Carlo estimates of any desired statistics. The complete CMS sampling procedure is detailed in Appendix~\ref{appendix:cms_sampling}.

\section{Experiments}
\label{sec:experiments}

\subsection{Experimental Setup}
We evaluate DeepLévy for multi-horizon probabilistic forecasting on three datasets: Bitcoin Returns, COVID-19 Cases, and Synthetic Lévy. We focus on three questions: overall probabilistic accuracy, tail-risk calibration, and the contribution of key design choices.


{\textbf{Datasets}.} We use Bitcoin Returns (hourly log-returns from Yahoo Finance~\citep{yahoofinance_btc}), COVID-19 Cases (daily JHU CSSE data~\citep{dong2020interactive}), and Synthetic Lévy sequences generated with the CMS method~\citep{chambers1976method}. Following Appendix~\ref{appendix:tail_diagnostics}, reported test metrics on the two real datasets are averaged over time segments pre-selected for pronounced heavy tails; they therefore reflect performance on volatile regimes rather than an unfiltered chronological test stream. Detailed preprocessing, split construction, and diagnostics are in Appendix~\ref{appendix:tail_diagnostics}.


\textbf{Baselines}. We compare against DeepAR variants (Gaussian, Laplace, Student's $t$, asymmetric Student's $t$, mixture Student's $t$, and mixture Gaussian), DeepVAR, DSSM, DeepFactor, and ProbTransformer, plus DeepLévy (Single Component, $K=1$) as an ablation. This set separates gains from skewness modeling, mixture flexibility, and explicit stable-tail parameterization. Full implementation details are in Appendix~\ref{appendix:implementation}.

\textbf{Evaluation Metrics}. We report CRPS~\citep{zamo2018estimation}, Tail-CRPS~\citep{wessel2025improving}, QL~\citep{steinwart2011estimating, koenker2005quantile}, Coverage (0.75/0.90/0.995), and PIT-KS~\citep{noceti2003evaluation}. Metrics are chosen to jointly evaluate global fit, extreme-tail performance, and calibration. Tail-CRPS as defined in~\citet{wessel2025improving} uses forecast-dependent integration limits; Appendix~\ref{appendix:metrics} records this and related caveats (properness, sample size for extreme coverage). Formal definitions are provided in Appendix~\ref{appendix:metrics}.

\subsection{Main Results}

Table~\ref{tab:main_results} shows a clear center-tail trade-off. Strong non-stable baselines (asymmetric and mixture Student's $t$) achieve the best CRPS/QL, while DeepLévy is best on tail metrics with Tail-CRPS 0.461 and Coverage@0.995 of 0.976 (deviation 0.019). The closest non-stable alternative (mixture Student's $t$) reaches Tail-CRPS 0.479 and Coverage@0.995 at 0.965. DeepAR with Gaussian or Laplace heads, DeepVAR, and DeepFactor all trail the stronger baselines by wide margins on CRPS/Tail-CRPS/QL and show larger extreme-quantile coverage deviations than Student's $t$-family and Transformer baselines, reflecting limitations of light-tailed or factorized likelihoods on our tail-heavy slices. This suggests that skewness and mixture flexibility help, but explicit stable-tail parameterization provides additional gains in extreme regions.

\begin{table*}[htbp]
\centering
\caption{Main results averaged across datasets/horizons (5 seeds). Best in \textbf{bold}; second-best values (including ties) are \underline{underlined}. $\downarrow$ indicates lower is better. Coverage is reported as value (absolute deviation from nominal).}
\label{tab:main_results}
\resizebox{\textwidth}{!}{
\begin{tabular}{@{}lccccccc@{}}
\toprule
Model & CRPS $\downarrow$ & Tail-CRPS $\downarrow$ & QL $\downarrow$ & PIT-KS $\downarrow$ & Cov@0.75 ($|\Delta|$) & Cov@0.90 ($|\Delta|$) & Cov@0.995 ($|\Delta|$) \\
\midrule
DeepAR (Gaussian) & 0.317 & 0.583 & 0.249 & \textbf{0.091} & 0.755 (0.005) & 0.861 (0.039) & 0.932 (0.063) \\
DeepAR (Laplace) & 0.302 & 0.551 & 0.233 & 0.108 & \underline{0.747} (0.003) & 0.872 (0.028) & 0.943 (0.052) \\
DeepAR (Student's $t$) & 0.271 & 0.502 & 0.198 & 0.123 & 0.734 (0.016) & 0.876 (0.024) & 0.954 (0.041) \\
DeepAR (Asym. Stud. $t$) & \underline{0.268} & 0.494 & \underline{0.194} & 0.114 & 0.742 (0.008) & 0.881 (0.019) & 0.961 (0.034) \\
DeepAR (Mix. Stud. $t$, $K=3$) & \textbf{0.266} & \underline{0.479} & \textbf{0.191} & 0.109 & 0.745 (0.005) & \underline{0.885} (0.015) & 0.965 (0.030) \\
DeepAR (Mix. Gaussian, $K=3$) & 0.286 & 0.528 & 0.214 & \underline{0.097} & \textbf{0.752} (0.002) & 0.878 (0.022) & 0.951 (0.044) \\
DeepVAR & 0.338 & 0.627 & 0.264 & 0.152 & 0.728 (0.022) & 0.853 (0.047) & 0.921 (0.074) \\
DSSM & 0.322 & 0.594 & 0.251 & 0.137 & 0.739 (0.011) & 0.867 (0.033) & 0.937 (0.058) \\
DeepFactor & 0.329 & 0.611 & 0.258 & 0.149 & 0.725 (0.025) & 0.851 (0.049) & 0.924 (0.071) \\
ProbTransformer & 0.298 & 0.537 & 0.231 & 0.099 & \textbf{0.752} (0.002) & 0.874 (0.026) & 0.948 (0.047) \\
\midrule
DeepL\'{e}vy (Single Component) & 0.301 & \underline{0.479} & 0.228 & 0.138 & 0.731 (0.019) & 0.884 (0.016) & \underline{0.969} (0.026) \\
\textbf{DeepL\'{e}vy} & 0.293 & \textbf{0.461} & 0.219 & 0.126 & 0.738 (0.012) & \textbf{0.889} (0.011) & \textbf{0.976} (0.019) \\
\bottomrule
\end{tabular}
}
\end{table*}

\begin{figure*}[ht]
\vskip 0.2in
\begin{center}
\centerline{\includegraphics[width=1\linewidth]{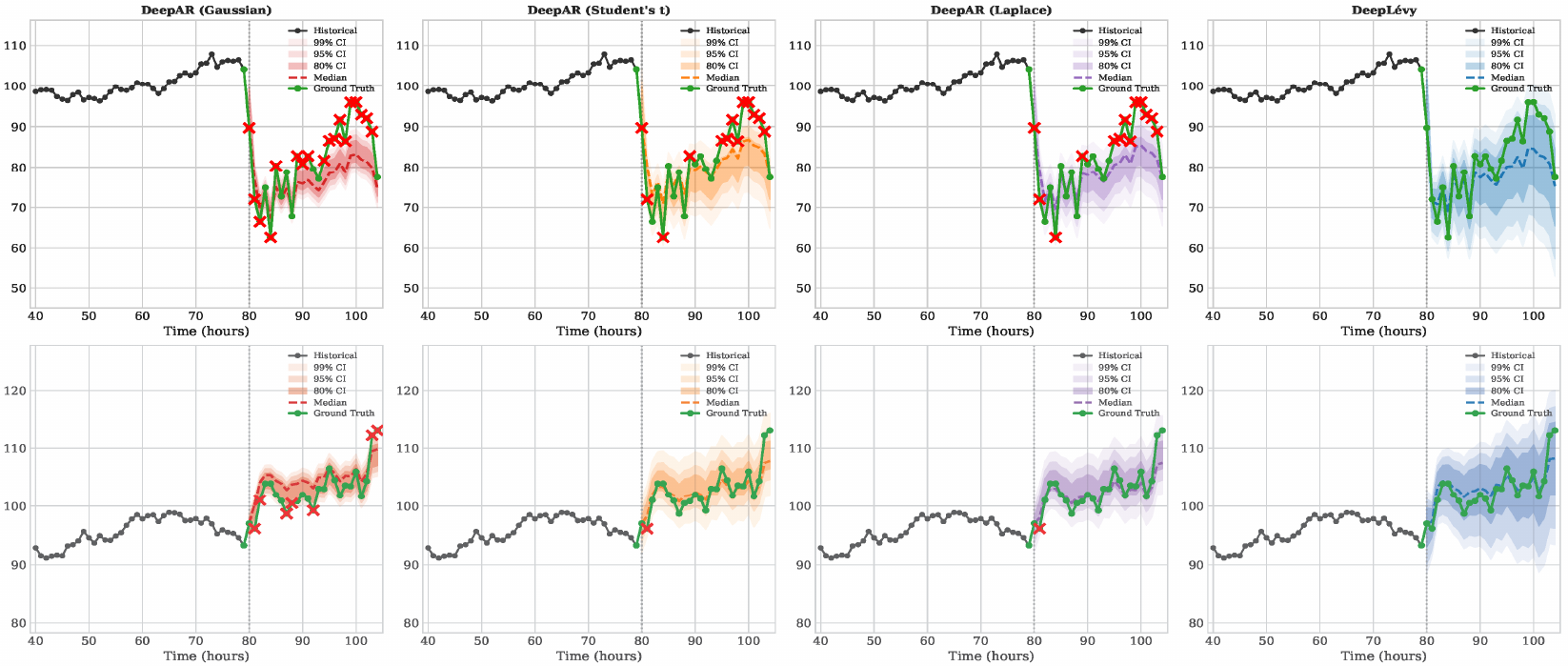}}
\caption{Forecast comparison on Bitcoin (filtered volatile segments; Appendix~\ref{appendix:tail_diagnostics}). Models are trained and scored on \textbf{preprocessed hourly log-returns}; the axis shows the \textbf{native scale of the plotted series} after inverse preprocessing used for visualization (so magnitudes are not raw log-returns). Red crosses ($\times$) mark ground-truth points outside the 99\% interval. DeepL\'{e}vy improves extreme-spike coverage.}
\label{fig:visual}
\end{center}
\vskip -0.2in
\end{figure*}

Table~\ref{tab:combined_results} reports per-dataset tail results. DeepLévy is best on all three datasets for both Tail-CRPS and Coverage@0.995. Among baselines, mixture and asymmetric Student's $t$ variants are the strongest runners-up, while DeepLévy (SC) is typically the closest model on Coverage@0.995. On every dataset column, DeepAR (Gaussian/Laplace), DeepVAR, and DeepFactor each exceed the next-worst remaining model on Tail-CRPS by more than $0.1$; DeepVAR and DeepFactor additionally register the largest Coverage@0.995 deviations, while Gaussian and Laplace heads still show substantially larger gaps than Student's $t$-family baselines though not always the single worst per column.
To quantify variability, all metrics in Tables~\ref{tab:main_results} and~\ref{tab:combined_results} are reported as means over 5 random seeds; complementary uncertainty intervals and compute-resource disclosure are provided in Appendix~\ref{appendix:repro_stat_compute}.

\begin{table}[t]
\centering
\caption{Per-dataset Tail-CRPS ($\downarrow$) and Coverage@0.995 (nominal 0.995). Coverage is shown as value (absolute deviation). Best in \textbf{bold}; second-best values are \underline{underlined}.}
\label{tab:combined_results}
\small
\setlength{\tabcolsep}{2.5pt} 
\begin{tabular}{@{}l ccc c ccc@{}}
\toprule
& \multicolumn{3}{c}{\textbf{Tail-CRPS ($\downarrow$)}} & & \multicolumn{3}{c}{\textbf{Coverage@0.995 (Value (Abs Dev))}} \\
\cmidrule{2-4} \cmidrule{6-8}
Model & Bitcoin & COVID-19 & Synth. & & Bitcoin & COVID-19 & Synth. \\
\midrule
DeepAR (Gaus.) & 0.594 & 0.637 & 0.518 & & 0.927 (0.068) & 0.916 (0.079) & 0.953 (0.042) \\
DeepAR (Lapl.) & 0.563 & 0.602 & 0.488 & & 0.938 (0.057) & 0.929 (0.066) & 0.962 (0.033) \\
DeepAR (Stud. $t$) & 0.509 & 0.558 & 0.439 & & 0.953 (0.042) & 0.941 (0.054) & 0.968 (0.027) \\
DeepAR (Asym. Stud. $t$) & 0.487 & \underline{0.536} & \underline{0.430} & & 0.961 (0.034) & 0.949 (0.046) & 0.971 (0.024) \\
DeepAR (Mix. Stud. $t$) & \underline{0.468} & 0.539 & 0.431 & & 0.968 (0.027) & 0.956 (0.039) & 0.973 (0.022) \\
DeepAR (Mix. Gaus.) & 0.532 & 0.579 & 0.459 & & 0.949 (0.046) & 0.938 (0.057) & 0.964 (0.031) \\
DeepVAR & 0.641 & 0.689 & 0.551 & & 0.917 (0.078) & 0.904 (0.091) & 0.942 (0.053) \\
DSSM & 0.608 & 0.652 & 0.522 & & 0.932 (0.063) & 0.922 (0.073) & 0.957 (0.038) \\
DeepFactor & 0.623 & 0.671 & 0.539 & & 0.921 (0.074) & 0.909 (0.086) & 0.946 (0.049) \\
ProbTrans. & 0.547 & 0.591 & 0.473 & & 0.946 (0.049) & 0.936 (0.059) & 0.963 (0.032) \\
\midrule
DeepL\'{e}vy (SC) & 0.467 & 0.536 & 0.434 & & \underline{0.971} (0.024) & \underline{0.960} (0.035) & \underline{0.976} (0.019) \\
\textbf{DeepL\'{e}vy} & \textbf{0.441} & \textbf{0.519} & \textbf{0.423} & & \textbf{0.979} (0.016) & \textbf{0.968} (0.027) & \textbf{0.981} (0.014) \\
\bottomrule
\end{tabular}
\end{table}

\subsection{Ablation Study}

Table~\ref{tab:ablation} examines the contribution of key components. The results reinforce the necessity of our architectural choices. Removing scale-adaptive weighting (Uniform Frequency) leads to a substantial degradation in Tail-CRPS (0.522 vs 0.441 on Bitcoin), confirming that high-frequency components must be re-weighted to match characteristic function norms. In the \emph{$S_1$ Parameterization} row, we replace only this CF map with the standard discontinuous $S_1$ parameterization (using the standard $\alpha=1$ branch), while keeping the rest of the architecture and training setup unchanged; therefore, the gap isolates the effect of parameterization choice. The $S_1$ variant performs significantly worse, particularly on COVID-19 (0.681 vs 0.519), consistent with instability around $\alpha=1$. The single-component model ($K=1$) lags behind the full mixture, particularly on complex real-world data (Bitcoin: 0.467 vs 0.441), validating the use of mixtures to capture multi-modal tail dynamics. Overall, results confirm a center-tail trade-off: asymmetric/mixture Student's $t$ variants are stronger on global metrics, while DeepLévy is strongest on extreme-tail risk. For applications prioritizing high-quantile reliability, the tail gains justify the modest central-mass trade-off.

\begin{table}[t]
\centering
\caption{Ablation on DeepL\'{e}vy components using Tail-CRPS ($\downarrow$). Best in \textbf{bold}.}
\label{tab:ablation}
\small
\setlength{\tabcolsep}{12pt} 
\begin{tabular}{@{}lccc@{}}
\toprule
Model Variant & Bitcoin & COVID-19 & Synthetic \\
\midrule
Full Model & \textbf{0.441} & \textbf{0.519} & \textbf{0.423} \\
Single Component ($K=1$) & 0.467 & 0.536 & 0.434 \\
Uniform Frequency Weighting & 0.522 & 0.585 & 0.463 \\
$S_1$ Parameterization & 0.563 & 0.681 & 0.547 \\
\bottomrule
\end{tabular}
\end{table}

\section{Conclusion}

We introduced DeepLévy, a neural framework for heavy-tailed probabilistic forecasting that models predictive distributions as mixtures of Lévy $\alpha$-stable components. By matching characteristic functions rather than maximizing intractable likelihoods, DeepLévy learns context-dependent tail behavior end-to-end via standard gradient descent. Experiments on financial and epidemiological time series demonstrate substantial improvements in tail risk quantification while remaining competitive on overall forecasting accuracy. Our work demonstrates the feasibility of integrating Lévy stable distributions with deep neural networks for heavy-tailed uncertainty learning. Current limitations include the parametric stable assumption and larger batch size requirements compared to likelihood-based methods. Future work will explore multivariate stable distributions and modeling temporal dependence in tail parameters.

\newpage
\bibliographystyle{plainnat}
\bibliography{reference}

\newpage
\appendix
\section{Additional Methodology Details}

\subsection{Numerical Stability Considerations for Parameter Constraints}
\label{appendix:numerical_stability}

This section explains why we use bounded parameter projections in Section~\ref{sec:projection}.

\textbf{Stability Index $\alpha$ Constraints.} The stability index is defined on $(0,2]$, but boundary values present significant challenges when using a fixed characteristic function (CF) grid and finite mini-batches. Specifically, as $\alpha$ approaches $0$, the distribution tails become extremely heavy and the CF decay becomes very slow, necessitating a much larger grid to capture the signal. Conversely, as $\alpha$ approaches $2$, the law nears a Gaussian distribution, causing the CF to shrink toward machine precision on a bounded grid, which weakens the numerical identifiability during the CF matching process. We therefore project $\hat{\alpha}_k^{(h)}$ to $[0.1,1.95]$ through Eq.~\eqref{eq:param_constraints}. This upper bound is an engineering choice for our grid and batch settings. It is not caused by a singularity at $\alpha=2$. One implication is that exact Gaussian behavior is not representable.

\textbf{Skewness Parameter $\beta$ Constraints.} The skewness parameter lies in $[-1,1]$. Values close to $\pm 1$ are numerically fragile in both CF evaluation and CMS sampling. We therefore keep predictions away from the boundary with $\hat{\beta}_k^{(h)}=(1-\epsilon_{\beta}^{\max})\tanh(\cdot)$. In our implementation, $\epsilon_{\beta}^{\max}=0.02$ and the effective range is $(-0.98,0.98)$. Sensitivity results are reported in Appendix~\ref{appendix:beta_sensitivity}.

\textbf{Scale Parameter $\gamma$ Constraints.} The scale must be strictly positive. We set $\epsilon_\gamma=10^{-4}$ as a safety floor. Very small scales can destabilize CF evaluation at large frequencies.

\begin{center}
\begin{minipage}{0.65\linewidth}
\begin{algorithm}[H]
\caption{Numerically Stable CF Computation under $S_0$ Parameterization}
\label{alg:cf_computation}
\begin{algorithmic}[1]
    \REQUIRE Frequency $\tau$, Parameters $(\alpha, \beta, \gamma, \delta)$
    \ENSURE Characteristic function value $\phi(\tau)$
    
    \IF{$|\tau| < 10^{-12}$}
        \STATE \textbf{return} $1$
    \ENDIF
    
    \STATE $s \leftarrow \text{sgn}(\tau)$
    \STATE $u \leftarrow |\gamma\tau|$
    \STATE $\psi_{\text{Re}} \leftarrow -u^\alpha$
    \STATE $g \leftarrow \mathrm{expm1}\bigl((1-\alpha)\ln(u + 10^{-10})\bigr)$ \COMMENT{$|\gamma\tau|^{1-\alpha}-1$ stably}
    \STATE $\psi_{\text{Im}} \leftarrow \delta\tau - u^\alpha \beta s \tan\left(\frac{\pi\alpha}{2}\right) g$
    
    \STATE $\psi_{\text{Re}} \leftarrow \max(\psi_{\text{Re}}, -50)$ \COMMENT{Prevent underflow}
    \STATE $\phi(\tau) \leftarrow \exp(\psi_{\text{Re}}) \cdot [\cos(\psi_{\text{Im}}) + i\sin(\psi_{\text{Im}})]$
    \STATE \textbf{return} $\phi(\tau)$
\end{algorithmic}
\end{algorithm}
\end{minipage}
\end{center}

\subsection{Detailed Characteristic Function Computation}
\label{appendix:cf_computation}

This section summarizes the numerically stable CF computation used in Section~\ref{sec:cf_matching}.

\textbf{Frequency Grid Construction.} The CF is defined on all real frequencies. In implementation we use a fixed symmetric grid:
\begin{equation}
\label{eq:freq_grid_appendix}
    \mathcal{T} = \left\{ \tau_m : \tau_m = -\tau_{\max} + \frac{2\tau_{\max}(m-1)}{M-1}, \; m = 1, \ldots, M \right\},
\end{equation}
where $M$ is odd so that zero is included in the grid. The parameter $\tau_{\max}$ controls frequency coverage. For standardized data we use $\tau_{\max}=15.0$. This captures informative frequencies while avoiding regions where the CF is already negligible.

\textbf{Log-Characteristic Function Computation.} Direct evaluation of $\varphi(\tau)$ can overflow or underflow. We therefore compute in log space and separate real and imaginary parts.

Let $u = |\gamma\tau|$ and $s = \text{sgn}(\tau)$. Under $S_0$, the log-characteristic function $\psi(\tau) = \ln\varphi(\tau) = \psi_{\text{Re}}(\tau) + i\,\psi_{\text{Im}}(\tau)$ is given for all $\alpha \neq 1$ by
\begin{equation}
\label{eq:log_cf_real_appendix}
    \psi_{\text{Re}}(\tau) = -u^\alpha,
\end{equation}
\begin{equation}
\label{eq:log_cf_imag_appendix}
    \psi_{\text{Im}}(\tau) = \delta\tau - u^\alpha \beta s \tan\frac{\pi\alpha}{2}\left(u^{1-\alpha} - 1\right),
\end{equation}
This form is continuous at $\alpha=1$ and matches the closed $\alpha=1$ expression in the limit. In code we use an \texttt{expm1}-based evaluation for numerical robustness near $\alpha=1$.

To prevent arithmetic overflow when exponentiating $\psi$, we apply value clipping to the real part:
\begin{equation}
\label{eq:clipping_appendix}
    \psi_{\text{Re}}(\tau) \leftarrow \max(\psi_{\text{Re}}(\tau), -50).
\end{equation}
The clipping level keeps numerical values stable while preserving useful signal.

The component CF is then computed using Euler's formula:
\begin{equation}
\label{eq:cf_component_appendix}
    \varphi_k^{(h)}(\tau) = \exp(\psi_{\text{Re}}(\tau)) \cdot \left[\cos(\psi_{\text{Im}}(\tau)) + i\sin(\psi_{\text{Im}}(\tau))\right],
\end{equation}
For each component and horizon, these expressions use the corresponding predicted parameters.

\subsection{Summary of Numerical Constants}
\label{appendix:numerical_constants}

Table~\ref{tab:numerical_constants} summarizes all numerical constants used in DeepLévy and their purposes.

\begin{table*}[ht]
\centering
\caption{Summary of numerical constants in DeepLévy.}
\label{tab:numerical_constants}
\begin{tabular}{lll}
\toprule
\textbf{Constant} & \textbf{Value} & \textbf{Purpose} \\
\midrule
$\alpha_{\min}$ & $0.1$ & Lower bound for stability index projection \\
$\alpha_{\max}$ & $1.95$ & Upper bound for stability index projection \\
$\Delta_\alpha$ & $1.85$ & Projection span, $\alpha_{\max}-\alpha_{\min}$ \\
$\epsilon_{\beta}^{\max}$ & $0.02$ & Margin from $|\beta|=1$, so $\hat{\beta}\in(-0.98,0.98)$ \\
$\epsilon_\gamma$ & $10^{-4}$ & Lower bound for scale parameter $\hat{\gamma}_k^{(h)}$ \\
$\epsilon_\alpha$ & $0.01$ & Optional speed shortcut: use closed $\alpha{=}1$ CF when $|\alpha-1| \le \epsilon_\alpha$ \\
$\epsilon_{\ln}$ & $10^{-10}$ & Prevents $\ln(0)$ in Cauchy-like CF formula \\
$\epsilon_\tau$ & $10^{-12}$ & Threshold for excluding $\tau \approx 0$ from loss \\
$\epsilon_W$ & $10^{-8}$ & Prevents division by zero in loss normalization \\
$\epsilon_H$ & $10^{-8}$ & Prevents $\ln(0)$ in entropy computation \\
$\epsilon_\beta$ & $10^{-6}$ & Threshold for $\beta \approx 0$ in CMS sampling \\
Clipping threshold & $-50$ & Prevents underflow in $\exp(\psi_{\text{Re}})$ \\
$\tau_{\max}$ & $15.0$ & Frequency grid bound (for standardized data) \\
\bottomrule
\end{tabular}
\end{table*}

These constants play different roles. Larger values provide safe lower bounds. Smaller values prevent singular operations such as logarithm at zero.
\subsection{Chambers-Mallows-Stuck Sampling Algorithm}
\label{appendix:cms_sampling}

This section describes CMS sampling for the $S_0$ parameterization.

\textbf{Component Selection.} We first sample a component index $k^* \in \{1, \ldots, K\}$ from the categorical distribution defined by the mixing weights:
\begin{equation}
\label{eq:component_sampling_appendix}
    k^* \sim \text{Categorical}(\boldsymbol{\pi}^{(h)}), \quad \text{i.e.,} \quad \mathbb{P}(k^* = k) = \pi_k^{(h)}.
\end{equation}

\textbf{Random Variable Generation.} We first draw two independent random variables:
\begin{equation}
\label{eq:random_vars_appendix}
    V \sim \text{Uniform}\left(-\frac{\pi}{2}, \frac{\pi}{2}\right), \quad W \sim \text{Exponential}(1),
\end{equation}
Here $\text{Exponential}(1)$ is the standard exponential distribution.

\textbf{Case 1: $|\hat{\alpha}_{k^*}^{(h)} - 1| > \epsilon_\alpha$ (General stable distribution).} We compute intermediate quantities:
\begin{equation}
\label{eq:cms_intermediate_appendix}
    \zeta = \hat{\beta}_{k^*}^{(h)} \tan\frac{\pi\hat{\alpha}_{k^*}^{(h)}}{2}, \quad \xi = \frac{1}{\hat{\alpha}_{k^*}^{(h)}}\arctan(\zeta),
\end{equation}
and the standardized sample:
\begin{equation}
\label{eq:cms_standard_appendix}
    Z = (1 + \zeta^2)^{1/(2\hat{\alpha}_{k^*}^{(h)})} \cdot \frac{\sin(\hat{\alpha}_{k^*}^{(h)}(V + \xi))}{(\cos V)^{1/\hat{\alpha}_{k^*}^{(h)}}} \cdot \left(\frac{\cos(V - \hat{\alpha}_{k^*}^{(h)}(V + \xi))}{W}\right)^{(1-\hat{\alpha}_{k^*}^{(h)})/\hat{\alpha}_{k^*}^{(h)}}.
\end{equation}

\textbf{Case 2: $|\hat{\alpha}_{k^*}^{(h)} - 1| \leq \epsilon_\alpha$ (Cauchy-like distribution).} The sampling formula simplifies:
\begin{equation}
\label{eq:cms_cauchy_appendix}
    Z = 
    \begin{cases}
        \tan V, & \text{if } |\hat{\beta}_{k^*}^{(h)}| < \epsilon_\beta, \\[6pt]
        \frac{2}{\pi}\left[\left(\frac{\pi}{2} + \hat{\beta}_{k^*}^{(h)} V\right)\tan V - \hat{\beta}_{k^*}^{(h)}\ln\left(\frac{\frac{\pi}{2}\,W\cos V}{\frac{\pi}{2} + \hat{\beta}_{k^*}^{(h)} V}\right)\right], & \text{otherwise},
    \end{cases}
\end{equation}
The threshold $\epsilon_\beta=10^{-6}$ separates symmetric and skewed Cauchy cases. The symmetric case uses a simpler and more stable form.

\textbf{Final Transformation.} The sample is transformed using the scale and location parameters:
\begin{equation}
\label{eq:final_sample_appendix}
    \hat{y}_{T+h} = \hat{\gamma}_{k^*}^{(h)} Z + \hat{\delta}_{k^*}^{(h)}.
\end{equation}

This gives efficient sampling from the predictive stable distribution.

\begin{center}
\begin{minipage}{0.75\linewidth}
\begin{algorithm}[H]
\caption{Chambers-Mallows-Stuck Sampling under $S_0$ Parameterization}
\label{alg:cms_sampling}
\begin{algorithmic}[1]
    \REQUIRE Parameters $(\alpha, \beta, \gamma, \delta)$
    \ENSURE Sample $X \sim S_0(\alpha, \beta, \gamma, \delta)$
    
    \STATE Generate $V \sim \text{Uniform}(-\frac{\pi}{2}, \frac{\pi}{2})$
    \STATE Generate $W \sim \text{Exponential}(1)$
    
    \IF{$|\alpha - 1| > 0.01$}
        \STATE \COMMENT{Standard case: $\alpha \neq 1$}
        \STATE $\zeta \leftarrow \beta\tan\frac{\pi\alpha}{2}$
        \STATE $\xi \leftarrow \frac{1}{\alpha}\arctan(\zeta)$
        \STATE $Z \leftarrow (1+\zeta^2)^{1/(2\alpha)} \cdot \frac{\sin(\alpha(V+\xi))}{(\cos V)^{1/\alpha}} \cdot \left(\frac{\cos(V-\alpha(V+\xi))}{W}\right)^{\frac{1-\alpha}{\alpha}}$
    \ELSE
        \STATE \COMMENT{Cauchy-like case: $\alpha \approx 1$}
        \IF{$|\beta| < 10^{-6}$}
            \STATE $Z \leftarrow \tan V$ \COMMENT{Pure Cauchy}
        \ELSE
            \STATE $Z \leftarrow \frac{2}{\pi}\left[\left(\frac{\pi}{2}+\beta V\right)\tan V - \beta\ln\left(\frac{\frac{\pi}{2}\,W\cos V}{\frac{\pi}{2}+\beta V}\right)\right]$
        \ENDIF
    \ENDIF
    
    \STATE $X \leftarrow \gamma Z + \delta$ \COMMENT{Scale and shift}
    \STATE \textbf{return} $X$
\end{algorithmic}
\end{algorithm}
\end{minipage}
\end{center}

\section{Additional Experimental Details}

\subsection{Heavy-Tail Diagnostics}
\label{appendix:tail_diagnostics}

Table~\ref{tab:tail_diagnostics_appendix} reports tail diagnostics on real datasets. We keep segments with high kurtosis and low Hill index so evaluation focuses on volatile regimes. Statistics are averaged over selected segments. Both datasets show heavy tails. Gaussian fits are rejected by KS tests, while stable fits are not rejected.

\begin{table}[ht]
\centering
\caption{Heavy-tail diagnostics for real-world datasets, averaged across multiple selected time segments. Hill $\hat{\alpha}$ estimates the tail index (lower values indicate heavier tails). Kurtosis $> 3$ indicates heavier tails than Gaussian. KS-test p-values test against Gaussian (rejected) and Stable (not rejected) distributions.}
\label{tab:tail_diagnostics_appendix}
\begin{tabular}{@{}lcccc@{}}
\toprule
Dataset & Hill $\hat{\alpha}$ & Kurtosis & KS (Gaussian) & KS (Stable) \\
\midrule
Bitcoin Returns & 1.58 & 42.6 & $p < 0.001$ & $p = 0.187$ \\
COVID-19 Cases & 1.43 & 56.3 & $p < 0.001$ & $p = 0.143$ \\
\bottomrule
\end{tabular}
\end{table}

\subsection{Implementation Details}
\label{appendix:implementation}

For all models, context windows are 48 for Bitcoin, 21 for COVID-19, and 50 for synthetic data. Prediction horizons are 12, 14, and 20. DeepLévy uses three mixture components, grid size 128, and $\tau_{\max}=15.0$. The encoder is a two-layer LSTM with 128 hidden units. The decoder is a one-layer LSTM with the same hidden size. Training runs for 100 epochs with Adam, learning rate $5\times10^{-4}$, cosine annealing, and batch size 256. The entropy weight is 0.01. Data splits are 70, 15, and 15 percent. We use gradient clipping with max norm 1.0. At inference, each test case uses 100 sampled trajectories. This gives a fair common budget across models, but extreme coverage at level 0.995 has notable Monte Carlo noise. Larger sample counts are preferable when tail calibration is the primary target.

All DeepAR variants share the same encoder and decoder. They differ only in output heads. Gaussian predicts location and scale. Laplace predicts location and scale. Student's $t$ predicts location, scale, and degrees of freedom with a positivity constraint. Asymmetric Student's $t$ adds a bounded skewness term. Mixture baselines use three components with softmax weights. They are trained with the same optimizer, batch size, and early stopping protocol.

\subsection{Theoretical Justification of CF Matching}
\label{appendix:cf_theory}

For a fixed context $\mathbf{x}_{1:T}$ and forecast horizon $h$, let $\phi^\star_h(\tau \mid \mathbf{x}_{1:T})$ denote the true conditional characteristic function (CF) of $Y_{T+h}$, and $\phi_{\theta,h}(\tau \mid \mathbf{x}_{1:T})$ the model CF induced by DeepL\'{e}vy. Consider the population weighted CF discrepancy
\begin{equation}
\label{eq:cf_pop_loss}
\mathcal{L}_{\mathrm{CF}}^{\mathrm{pop}}(\theta \mid \mathbf{x}_{1:T})
= \int_{\mathbb{R}} w_h(\tau)\,
\left|\phi_{\theta,h}(\tau \mid \mathbf{x}_{1:T})-\phi^\star_h(\tau \mid \mathbf{x}_{1:T})\right|^2\, d\tau,
\end{equation}
with measurable $w_h(\tau)\ge 0$ and positive weight for almost every $\tau$ on $\mathbb{R}$.

\textbf{Proposition 1 (Population identification).}
If $\mathcal{L}_{\mathrm{CF}}^{\mathrm{pop}}(\theta \mid \mathbf{x}_{1:T})=0$, then
$\phi_{\theta,h}(\tau \mid \mathbf{x}_{1:T})=\phi^\star_h(\tau \mid \mathbf{x}_{1:T})$ for almost every $\tau$, and therefore the model and target conditional distributions are equal:
\begin{equation}
P_{\theta,h}(\cdot \mid \mathbf{x}_{1:T}) = P^\star_h(\cdot \mid \mathbf{x}_{1:T}).
\end{equation}

\textit{Proof sketch.}
The integrand in Eq.~\eqref{eq:cf_pop_loss} is nonnegative. Zero loss implies equality of model and target CF for almost every frequency. CFs are uniformly continuous, so almost-everywhere equality implies pointwise equality. The uniqueness theorem for characteristic functions then gives equality in distribution. \hfill $\square$

\textbf{Lemma 1 (Empirical CF moments).}
For i.i.d. samples $\{Y_b\}_{b=1}^B \sim P_h^\star(\cdot \mid \mathbf{x}_{1:T})$, define
\begin{equation}
\widehat{\phi}_{B,h}(\tau \mid \mathbf{x}_{1:T})=\frac{1}{B}\sum_{b=1}^{B} e^{i\tau Y_b}.
\end{equation}
Then, for each fixed $\tau$,
\begin{equation}
\mathbb{E}\left[\widehat{\phi}_{B,h}(\tau \mid \mathbf{x}_{1:T})\right]
=\phi_h^\star(\tau \mid \mathbf{x}_{1:T}), \quad
\mathrm{Var}\!\left(\widehat{\phi}_{B,h}(\tau \mid \mathbf{x}_{1:T})\right)\le \frac{1}{B}.
\end{equation}
Moreover,
\begin{equation}
\mathrm{Var}\!\left(\widehat{\phi}_{B,h}\right)
= \frac{1}{B}\mathrm{Var}\!\left(e^{i\tau Y}\right)
\le \frac{1}{B}\mathbb{E}\!\left[\left|e^{i\tau Y}\right|^2\right]
= \frac{1}{B},
\end{equation}
since $\left|e^{i\tau Y}\right|=1$. Estimation noise therefore decays as $O(B^{-1/2})$. This supports moderately large mini-batches. \textbf{Remark.} Lemma~1 studies repeated samples at one fixed context. Training in Eq.~\eqref{eq:cf_loss} averages across different contexts in each mini-batch. The lemma is used as a variance intuition, not as an exact decomposition of the training objective.

\subsection{Reproducibility}
\label{appendix:repro_stat_compute}

All reported results use five independent seeds with fixed data splits. For Tail-CRPS we also report standard deviation and 95 percent confidence intervals in Table~\ref{tab:stat_compute_appendix}. Most variation comes from random initialization and mini-batch order.

Table~\ref{tab:stat_compute_appendix} also reports hardware and representative runtime. Experiments ran on institutional servers with one NVIDIA V100 with 32GB memory per run, Xeon-class CPU, and 256GB RAM. The total budget for final reported runs is about 1,200 GPU-hours.

\textbf{Code availability.} An anonymized implementation with environment setup and run instructions is provided at \url{https://anonymous.4open.science/status/DeepLevy-AD1E}.

\begin{table}[ht]
\centering
\caption{Statistical significance and compute disclosure for representative models (Bitcoin benchmark). Tail-CRPS is mean $\pm$ std over 5 seeds, with 95\% CI in brackets.}
\label{tab:stat_compute_appendix}
\small
\begin{tabular}{@{}lccc@{}}
\toprule
Model & Tail-CRPS & Train Time (min/epoch) & Inference (ms/sample) \\
\midrule
DeepAR (Student's $t$) & $0.509 \pm 0.012$ [$0.498$, $0.520$] & 0.24 & 0.9 \\
DeepAR (Mix. Stud. $t$) & $0.468 \pm 0.010$ [$0.459$, $0.477$] & 0.29 & 1.1 \\
DeepL\'{e}vy & $\mathbf{0.441 \pm 0.009}$ [$0.433$, $0.449$] & 0.33 & 1.3 \\
\bottomrule
\end{tabular}
\end{table}

\subsection{Broader Impact and Asset Licenses}
\label{appendix:impact_license}

\textbf{Broader impact.} This work can improve reliability of extreme-event forecasting in financial risk and public health planning. Better tail calibration can reduce under-preparation for rare shocks. Risks remain. Users may over-trust automated scores, react to false alarms, or miss shifts outside the modeled regime. We recommend human oversight, routine calibration checks, and periodic retraining.

\textbf{Assets and licenses.} We use public datasets and public baseline descriptions. Bitcoin returns come from Yahoo Finance under provider terms. COVID-19 case data come from the Johns Hopkins CSSE repository under CC BY 4.0. Baselines are reimplemented from original papers or official repositories with citation. No closed third-party model weights are redistributed.

\subsection{Detailed Metric Definitions}
\label{appendix:metrics}

\textbf{Continuous Ranked Probability Score (CRPS)} measures the overall quality of probabilistic forecasts:
\begin{equation}
    \text{CRPS}(\hat{F}, y) = \int_{-\infty}^{\infty} \left(\hat{F}(x) - \mathbb{1}(y \leq x)\right)^2 dx.
\end{equation}

\textbf{Tail-weighted CRPS (Tail-CRPS)} focuses on lower and upper tails using the 10th and 90th forecast percentiles~\citep{wessel2025improving}:
\begin{equation}
\label{eq:tail_crps}
    \text{Tail-CRPS}(\hat{F}, y) = \int_{-\infty}^{q_{0.1}} \left(\hat{F}(x) - \mathbb{1}(y \leq x)\right)^2 dx + \int_{q_{0.9}}^{\infty} \left(\hat{F}(x) - \mathbb{1}(y \leq x)\right)^2 dx,
\end{equation}
where $q_{0.1}$ and $q_{0.9}$ are forecast percentiles. This metric is forecast dependent and is not strictly proper in the sense of~\citet{gneiting2014probabilistic,gneiting2011comparing}. A model can reduce penalty by widening dispersion. We still report it because all methods are scored under the same rule, which makes relative comparison useful for tail behavior. A stricter alternative is to use fixed external thresholds.

\textbf{Quantile Loss (QL)} evaluates the accuracy of specific quantile predictions with asymmetric penalty. We report the average across quantiles $\tau \in \{0.1, 0.5, 0.9, 0.99\}$.

\textbf{Coverage} at level $\tau$ measures the empirical frequency that observations fall below the predicted $\tau$-quantile. Well-calibrated models should achieve coverage close to the nominal level.

\textbf{PIT-KS} quantifies calibration via the Kolmogorov-Smirnov statistic on Probability Integral Transform values.

\subsection{Sensitivity to Skewness-Bound Margin}
\label{appendix:beta_sensitivity}

We test sensitivity to the skewness margin with three settings. All other training settings follow Appendix~\ref{appendix:implementation}. Results are means over five seeds on the Bitcoin validation split.

\begin{table}[ht]
\centering
\caption{Sensitivity to the skewness-bound margin $\epsilon_{\beta}^{\max}$. Lower Tail-CRPS is better.}
\label{tab:beta_sensitivity}
\begin{tabular}{lccc}
\toprule
$\epsilon_{\beta}^{\max}$ & $\max|\hat{\beta}|$ & Tail-CRPS $\downarrow$ & Unstable runs (\%) \\
\midrule
0.05 & 0.95 & 0.447 & 0.0 \\
0.02 & 0.98 & \textbf{0.441} & 0.0 \\
0.01 & 0.99 & 0.444 & 20.0 \\
\bottomrule
\end{tabular}
\end{table}
\textbf{Aggregation of unstable runs.} Tail-CRPS is averaged over runs that complete training. Runs marked as unstable are excluded. When instability is nonzero, the reported mean can be optimistic.

\subsection{Performance Across Tail Heaviness Regimes}
\label{appendix:tail_regimes}

To study performance across tail regimes, we split filtered Bitcoin test observations into three buckets using a rolling Hill estimate. Buckets correspond to low, medium, and high volatility. Table~\ref{tab:tail_heaviness_appendix} reports results in each bucket. A similar pattern is observed on COVID-19.

\begin{table*}[htbp]
\centering
\caption{Performance across tail heaviness regimes on \textbf{Bitcoin} filtered test points, partitioned by local Hill $\hat{\alpha}$: Low Volatility ($\hat{\alpha} \approx 1.8$, near-Gaussian), Medium Volatility ($\hat{\alpha} \approx 1.4$), and High Volatility ($\hat{\alpha} \approx 1.0$, extreme heavy-tail).}
\label{tab:tail_heaviness_appendix}
\resizebox{\textwidth}{!}{
\begin{tabular}{@{}l|ccc|ccc@{}}
\toprule
& \multicolumn{3}{c|}{Tail-CRPS $\downarrow$} & \multicolumn{3}{c}{Coverage@0.995 (deviation from 0.995)} \\
Model & Low Vol. ($\hat{\alpha}{\approx}1.8$) & Med Vol. ($\hat{\alpha}{\approx}1.4$) & High Vol. ($\hat{\alpha}{\approx}1.0$) & Low Vol. & Med Vol. & High Vol. \\
\midrule
DeepAR (Gaussian) & 0.662 & 0.792 & 0.982 & 0.761 (0.234) & 0.735 (0.260) & 0.688 (0.307) \\
DeepAR (Laplace) & 0.651 & 0.805 & 0.971 & 0.752 (0.243) & 0.721 (0.274) & 0.674 (0.321) \\
DeepAR (Student's $t$) & \textbf{0.398} & \underline{0.496} & 0.631 & \textbf{0.974} (0.021) & \underline{0.958} (0.037) & 0.928 (0.067) \\
DeepVAR & 0.772 & 0.902 & 1.082 & 0.661 (0.334) & 0.635 (0.360) & 0.588 (0.407) \\
DSSM & 0.449 & 0.579 & 0.768 & 0.962 (0.033) & 0.935 (0.060) & 0.888 (0.107) \\
DeepFactor & 0.765 & 0.895 & 1.069 & 0.660 (0.335) & 0.634 (0.361) & 0.587 (0.408) \\
ProbTransformer & \underline{0.408} & 0.531 & 0.698 & \underline{0.972} (0.023) & 0.951 (0.044) & 0.912 (0.083) \\
\midrule
DeepLévy Single Component& 0.421 & 0.502 & \underline{0.608} & 0.969 (0.026) & 0.956 (0.039) & \underline{0.941} (0.054) \\
\textbf{DeepLévy} & 0.412 & \textbf{0.468} & \textbf{0.551} & 0.971 (0.024) & \textbf{0.964} (0.031) & \textbf{0.954} (0.041) \\
\bottomrule
\end{tabular}
}
\end{table*}

\textbf{Key Observations.} In low volatility, Student's $t$ performs best and data is close to near Gaussian behavior. In medium volatility, DeepLévy starts to outperform Student's $t$. In high volatility, the DeepLévy advantage becomes larger on both Tail-CRPS and extreme coverage. Across buckets, DeepAR (Gaussian/Laplace), DeepVAR, and DeepFactor remain more than $0.1$ worse than the next-worst baseline on Tail-CRPS; DeepVAR and DeepFactor show the largest Coverage@0.995 gaps in each bucket, with Gaussian and Laplace still far behind Student's $t$-family models on extreme coverage.

\newpage
\subsection{Multi-Horizon Performance and Tail Calibration Analysis}
\label{appendix:horizon_calibration}

\begin{figure}[ht]
\centering
\includegraphics[width=0.8\columnwidth]{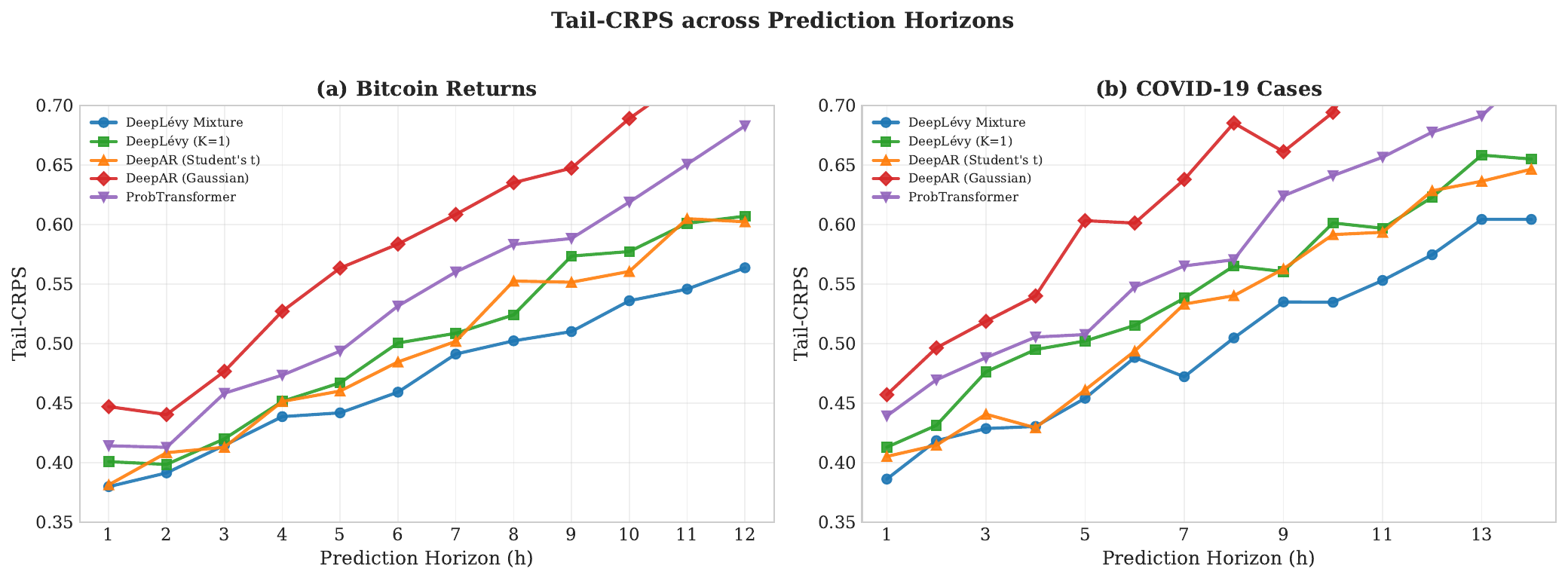}
\vspace{-0.3cm}
\caption{Tail-CRPS across prediction horizons on filtered test segments. \textbf{Left:} Bitcoin. \textbf{Right:} COVID-19. DeepL\'{e}vy (mixture) achieves lower Tail-CRPS than DeepAR (Student's $t$) across the displayed horizons on both panels, with the gap tending to widen as $h$ increases and multi-step uncertainty accumulates.}
\label{fig:horizon_analysis_appendix}
\end{figure}

\begin{figure}[ht]
\centering
\includegraphics[width=0.8\columnwidth]{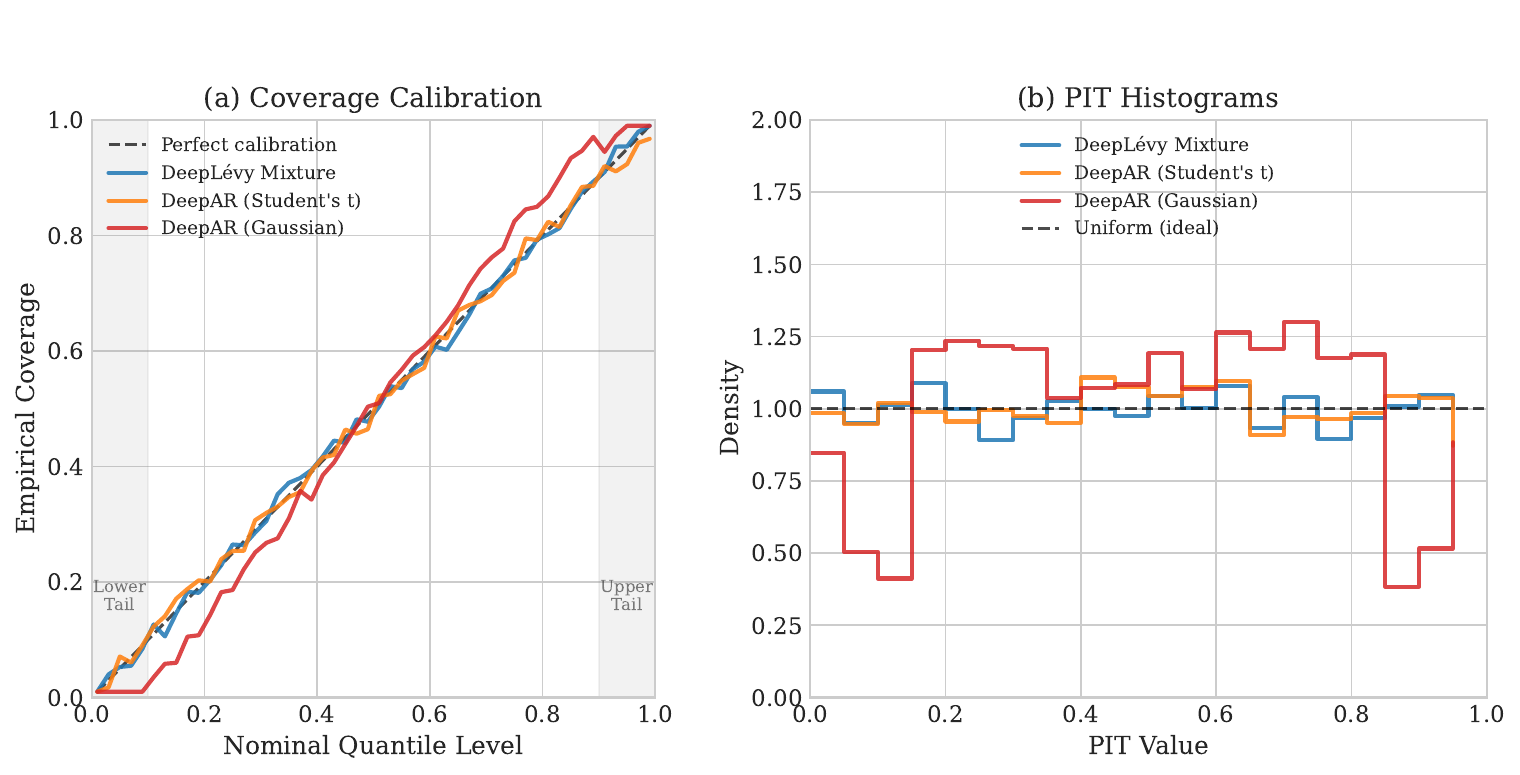}
\caption{Calibration analysis  \textbf{Left:} Empirical coverage vs.\ nominal quantile level; deviations at the corners indicate tail miscalibration. \textbf{Right:} PIT histograms; U-shaped mass for DeepAR with Gaussian or Laplace heads signals tail underestimation (overconfidence), whereas DeepLévy shows a more uniform profile.}
\label{fig:calibration_appendix}
\end{figure}

Figure~\ref{fig:horizon_analysis_appendix} shows how tail performance changes across prediction horizons on Bitcoin and COVID-19. Figure~\ref{fig:calibration_appendix} shows calibration through coverage curves and PIT histograms.

\vspace{-0.3cm}

\newpage
\subsection{Show Case Visualisation}
These panels visualize tail coverage under volatile periods. They show that DeepLévy captures both upward spikes and downward drops. Axes are shown in visualization scale after inverse preprocessing. Training and metrics use preprocessed targets from Appendix~\ref{appendix:tail_diagnostics}.

\vspace{-0.3cm}
\begin{figure*}[!ht]
\vskip 0.2in
\begin{center}
\centerline{\includegraphics[width=0.98\linewidth]{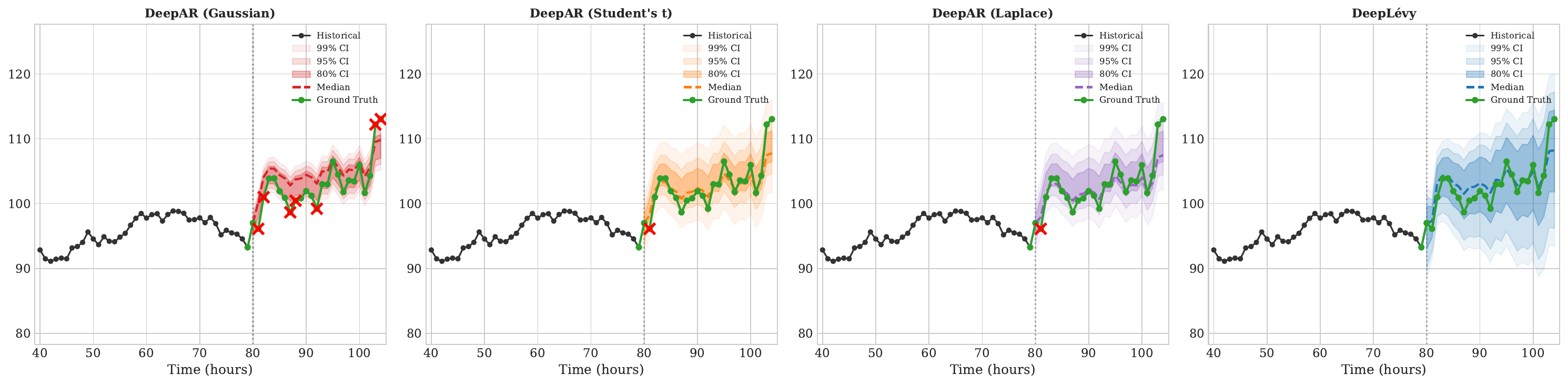}}
\label{fig:visual1}
\end{center}
\vskip -0.2in
\end{figure*}

\vspace{-0.3cm}
\begin{figure*}[ht]
\vskip 0.2in
\begin{center}
\centerline{\includegraphics[width=0.98\linewidth]{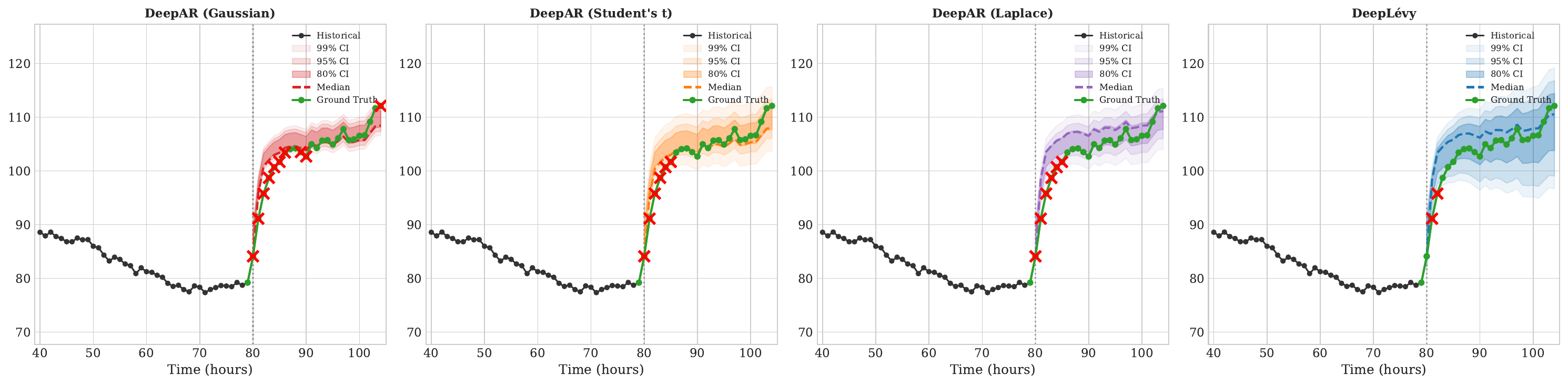}}
\label{fig:visual2}
\end{center}
\vskip -0.2in
\end{figure*}

\vspace{-0.3cm}
\begin{figure*}[ht]
\vskip 0.2in
\begin{center}
\centerline{\includegraphics[width=0.98\linewidth]{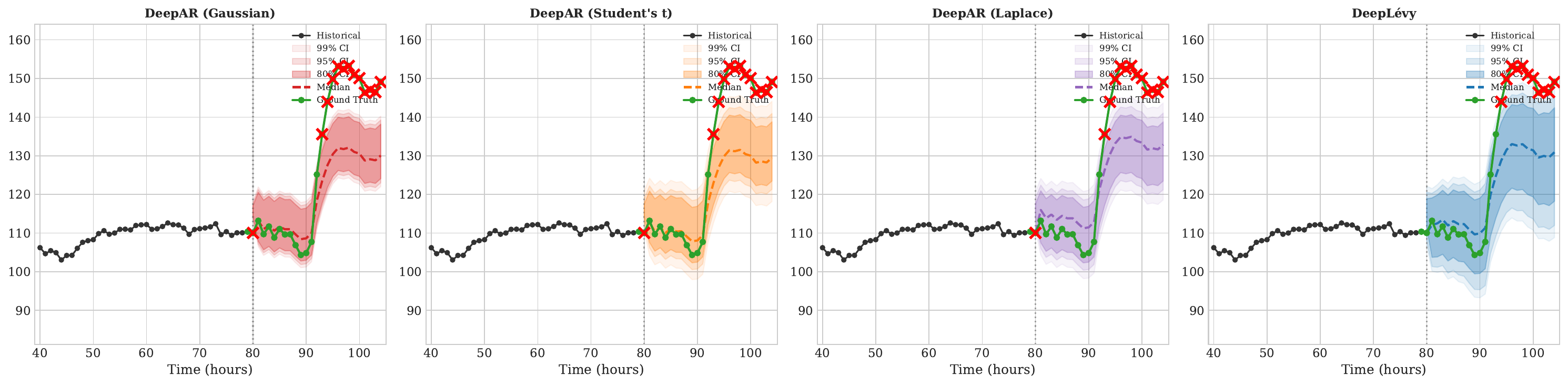}}
\label{fig:visual3}
\end{center}
\vskip -0.2in
\end{figure*}

\vspace{-0.3cm}
\begin{figure*}[!h]
\vskip 0.2in
\begin{center}
\centerline{\includegraphics[width=0.98\linewidth]{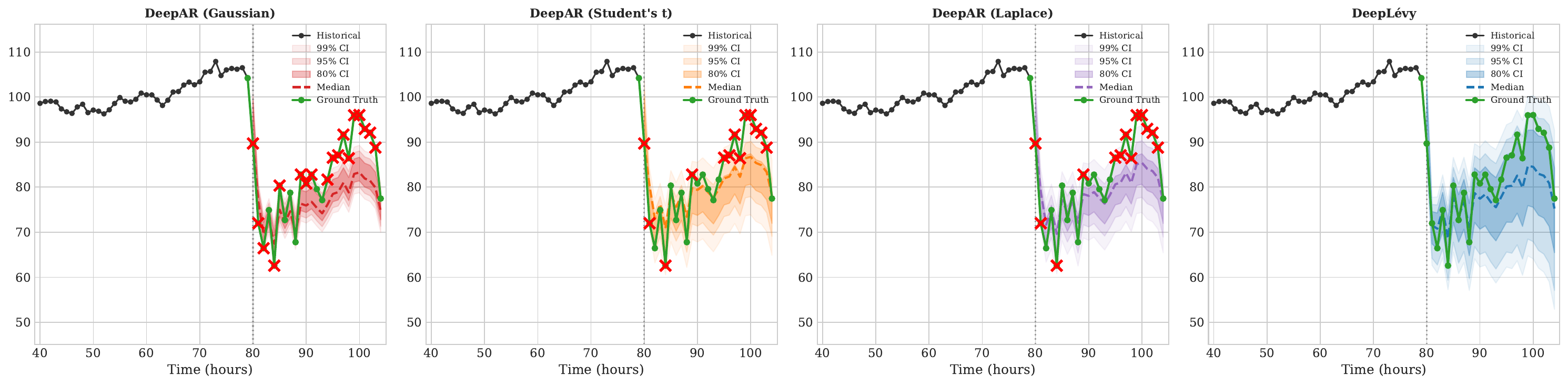}}
\label{fig:visual4}
\end{center}
\vskip -0.2in
\end{figure*}

\end{document}